\pdfoutput=1

\documentclass[11pt]{article}

\usepackage{acl}

\usepackage{times}
\usepackage{latexsym}

\usepackage[T1]{fontenc}

\usepackage[utf8]{inputenc}

\usepackage{microtype}
\usepackage{pifont}
\usepackage{float}
\usepackage{tabularx}
\usepackage{todonotes}
\usepackage{booktabs, multirow} 
\usepackage{soul}
\usepackage[table]{colortbl} 
\usepackage{changepage,threeparttable} 
\usepackage{siunitx}
\usepackage{dcolumn}
\definecolor{bettergreen}{RGB}{52, 168, 63}
\definecolor{betterred}{RGB}{234, 68, 63}

\definecolor{hlblue}{RGB}{173,216,230}
\definecolor{hlred}{RGB}{224,102,102}

\newcommand{\cmark}{\textcolor{bettergreen}{\ding{51}}}%
\newcommand{\xmark}{\textcolor{betterred}{\ding{55}}}%
\title{Subword-Delimited Downsampling for Better Character-Level Translation}

\author{Lukas Edman \qquad  Antonio Toral \qquad Gertjan van Noord \vspace{.2cm}
 \\ Center for Language and Cognition \\ 
 University of Groningen \vspace{.1cm}
 \\ {\tt \small\{j.l.edman, a.toral.ruiz, g.j.m.van.noord\}@rug.nl}
}
\begin{document}
\maketitle
\begin{abstract}
Subword-level models have been the dominant paradigm in NLP. However, character-level models have the benefit of seeing each character individually, providing the model with more detailed information that ultimately could lead to better models. Recent works have shown character-level models to be competitive with subword models, but costly in terms of time and computation. Character-level models with a downsampling component alleviate this, but at the cost of quality, particularly for machine translation. 
This work analyzes the problems of previous downsampling methods and introduces a novel downsampling method which is informed by subwords.
This new downsampling method not only outperforms existing downsampling methods, showing that downsampling characters can be done without sacrificing quality, but also leads to promising performance compared to subword models for translation.
\end{abstract}

\section{Introduction}
Character-level models (henceforth character models) have recently sparked interest in their potential applicability across a wide range of NLP tasks. They promise a tokenization-free approach, while also potentially allowing the model to quickly recognize similarities between words based on their spelling. However, as of yet, character models have not stood up to the level of subword models, mainly due to their similar performance while being significantly slower and more expensive to train due to their longer input sequences \cite{xue2022byt5}. 

To alleviate the problem of training time, several methods have been proposed to initially downsample characters into shorter sequences, which are then fed into the encoder or decoder. For discriminative tasks, these can be applied without any loss in performance \cite{tay2021charformer}, however for generative tasks like NMT, the performance is either untested or lacking when compared to character models without downsampling \cite{libovicky2021don}.


Seeing as subword tokenization is essentially a form of downsampling and performs quite well, the idea of downsampling is not inherently flawed. However, attempts to downsample within the neural network have not achieved similar performance for translation. This begs the question, why are current neural downsampling methods underperforming for translation?


In this work, we make three main contributions: 
\begin{enumerate}
    \setlength\itemsep{-0.2em}
    \item We analyze the existing downsampling methods based on their position, length, and morpheme consistency.
    \item We introduce a novel neural downsampling method based on subwords that outperforms existing downsampling methods.
    \item We make the necessary modifications to allow for variable-length downsampling and upsampling in an encoder-decoder architecture.
\end{enumerate}




\begin{table*}[!htp]\centering
\scriptsize
\begin{tabular}{lrrrrrrr}\toprule
 &\multicolumn{2}{c}{Downsampling Method} &\multicolumn{4}{c}{Architecture} \\ \cmidrule(lr){2-3} \cmidrule(lr){4-7}
\multicolumn{1}{l}{Work} &Encoder &Decoder &Downsampler &Encoder &Decoder &Upsampler \\ \midrule
\citet{lee2017fully} &Fixed &--- &CNN &LSTM &LSTM &LSTM \\
\citet{boukkouri2020characterbert} &Word &--- &CNN &Transformer &--- &--- \\
\citet{tay2021charformer} &Fixed &--- &GBST &Transformer &Transformer &--- \\
\citet{libovicky2021don} &Fixed &Fixed &CNN &Transformer &Transformer &LSTM \\
\citet{xue2022byt5} &--- &--- &--- &Transformer &Transformer &--- \\
\citet{edman2022patching} &Fixed &Fixed &GBST &Transformer &Transformer &LSTM \\
Ours &Subword &Subword &CNN &Transformer &Transformer &LSTM \\
\bottomrule
\end{tabular}
\caption{Summary of previous work on character-level models compared to ours. Note that many of the works have tested multiple models, so we only include their main additions which are unique in research.}\label{tab:prev_work}
\end{table*}

We start by providing an overview of the prior work in character-level NLP in Section~\ref{sect:old_work}. We then compare the various downsampling methods, noting the 3 main advantages of downsampling based on subwords in Section~\ref{sect:downsampling}. Next, we cover the modifications necessary to do variable-length downsampling for translation in Section~\ref{sect:arch}. We follow this with experimental details (Section~\ref{sect:exp}), results (Section~\ref{sect:results}), analysis (Section~\ref{sect:analysis}), discussion (Section~\ref{sect:discussion}), conclusion (Section~\ref{sect:conc}), and finally the limitations (Section~\ref{sect:limit}).


\section{Related Work} \label{sect:old_work}

Character-level models have been of interest for several years, notably used for character-level translation prior to the advent of Transformer models with some success \cite{costa2016character, lee2017fully}.

\citet{lee2017fully} raises the issue that the length of the sequences require the models to potentially capture much longer range dependencies, and as such introduces a downsampling method. This method consists of convolutional layers followed by a sequence-length-wise max pooling. The convolutional layers serve to learn local patterns in the characters and the max pooling is intended to reduce the length of the input, alleviating the long-range dependency issue. Reducing the length with a downsampling method such as max pooling can be thought of as a transformation from character tokens to pseudo-word tokens.

More recently, character-level NLP has been investigated with the use of the Transformer. The Transformer, while better able to handle longer sequences than RNNs, can similarly suffer when on the character-level due to the $O(n^2)$ complexity of self-attention. Nevertheless, ByT5 \cite{xue2022byt5}, a multilingual unsupervised pretrained character-level model, has shown comparable results to its subword-level counterpart mT5, while demonstrating some beneficial properties such as robustness to character-level noise. It is however slower than subword models both in training and test time. 

The Charformer \cite{tay2021charformer} reintroduces downsampling using a novel downsampling method, GBST, which uses a learned, weighted-average of character n-grams for each downsampled token. This shows similar performance to ByT5 while also being faster, however its performance on generative tasks such as NMT appears less promising. \citet{edman2022patching} investigate the usefulness of Charformer's GBST method for NMT, finding that using GBST decoder-side does not work out-of-the-box due to an information leak, and that even with a fix to the leak, it does not perform up to the level of the aforementioned convolutional downsampling method. 

Similar to the Charformer, CharacterBERT \cite{boukkouri2020characterbert} shows that incorporating character information can be useful on encoder-only tasks. They use a CNN similar to \citet{lee2017fully}'s, but downsample based on the length of the whole word, rather than at a fixed size. Their results show better generalization than subword models on classification of medical data, despite it not seeing any medical data in pretraining. They attribute this to its more generalized internal vocabulary as a result of receiving characters as input. 

In the context of NMT, \citet{libovicky2021don} attempt to answer why the current state-of-the-art models are not character models, to which the answer appears that their performance is not superior to subword models, and that downsampling methods sacrifice quality for efficiency. 

In doing so, \citeauthor{libovicky2021don} convert existing character models such as \citet{lee2017fully}'s to the Transformer architecture. With this, they propose a two-step decoding method, which adds an LSTM layer that takes as input the hidden representation of the Transformer decoder, concatenated with separately-learned character embeddings. The light-weight nature of the two-step decoder means little computation time is added. 

We show a tabular summary of the relevant previous work in Table~\ref{tab:prev_work}.

\section{Exploring Downsampling Methods} \label{sect:downsampling}
\begin{table*}[!htp]\centering
\scriptsize
\begin{tabular}{llccc}\toprule
& & \multicolumn{3}{c}{Consistency} \\ \cmidrule(lr){3-5}
Downsampling Method &Example & Position & Length & Morpheme \\ \midrule
Fixed & \includegraphics[scale=0.2]{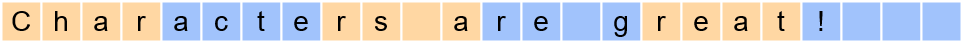} & \xmark & \cmark & \xmark \\ 
Buffered Fixed 
 & \includegraphics[scale=0.2]{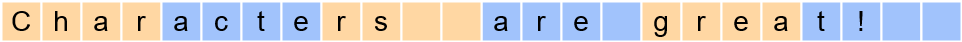} & \cmark & \cmark & \xmark \\ 
WDD & \includegraphics[scale=0.2]{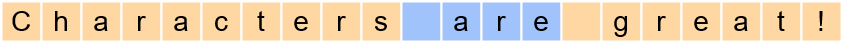} & \cmark & \xmark & \cmark \\ 
SDD & \includegraphics[scale=0.2]{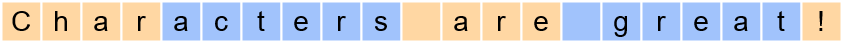} & \cmark & \cmark & \cmark \\
\bottomrule
\end{tabular}
\caption{The different downsampling methods tested. Alternating colors indicate the different downsampling blocks. In summary, Fixed and Buffered Fixed always downsample the same number of characters (in this case 4), with Buffered Fixed adding extra spaces between words so that each word begins at the beginning of a downsampling block. WDD downsamples based on words (defined by spaces), and SDD downsamples based on subwords, defined by a subword tokenizer.}\label{tab:ds_methods}
\end{table*}

We first compare our novel subword-based downsampling method over other possible forms of downsampling. Our choice of downsampling based on subwords is motivated by 3 factors:
\begin{enumerate}
    \setlength\itemsep{-0.2em}
    \item Positional consistency
    \item Length consistency
    \item Morpheme consistency
\end{enumerate}
We compare our subword-delimited downsampling (``SDD'') to the existing 2 methods, fixed-size downsampling (``Fixed'', used in \citet{lee2017fully} among others) and word-delimited downsampling (``WDD'', used in \citet{boukkouri2020characterbert}), as well as a third method, buffered fixed-size downsampling (``Buffered Fixed''), which we introduce to better understand the importance of position and morpheme consistency. 
Table~\ref{tab:ds_methods} shows an example of these downsampling methods.

\paragraph{Positional Consistency} 
The first factor we consider is the importance of positional consistency, that is, where a word begins within a downsampling block. For example, consider the following two sentences, with alternating colors denoting the chunking of character sequences when using a fixed-size downsampling factor of 4:
\begin{figure}[H]
    \centering
    \includegraphics[scale=0.205]{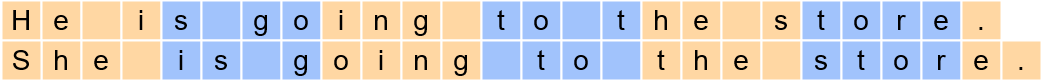}
\end{figure}
Words such as ``is'' or ``the'' end up in 2 tokens for some sentences, while only in 1 for others. Meanwhile, longer words such as ``going'' and ``store'' can be split several different ways, leading to several different potential representations depending on the sentence. This positional inconsistency introduces an extra level of difficulty to the model, which we expect results in worse performance.
Of the 4 methods tested, Fixed is the only one that suffers from this positional inconsistency. 

\paragraph{Length Consistency} \label{subsect:len_cons}
The consistency of the lengths of downsampling blocks is also important, particularly in the case of longer words. In the CNN downsampler, the max pooling acts as a bottleneck, making it more difficult for the model to learn a complete representation for words with many characters. 
For example, in Table~\ref{tab:ds_methods}, we see that the WDD method downsamples ``Characters'' into a single block, which means the max pooling downsamples that word by a factor of 10.
Furthermore, the LSTM in the upsampling module may have difficulty decoding a long sequence of characters from a single hidden representation.

This consistency mainly affects the WDD method. There is also a small amount of inconsistency in the SDD method, but this can be greatly minimized by setting a maximum subword token length (see details in Section~\ref{sec:upsampling}). 

\paragraph{Morpheme Consistency}
The third and final benefit of SDD is its creation of more morphologically consistent tokens. When splitting words into multiple subword tokens, it may be better for a model to split along the morpheme boundaries rather than every 4 characters, as the importance of characters can vary. Observe the effect of fixed-size splitting on various verbs: 
\begin{figure}[H]
    \centering
    \includegraphics[scale=0.205]{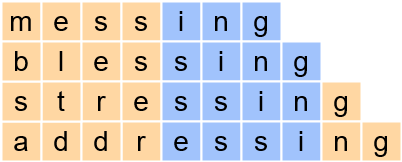}
\end{figure}

The ``ing'' ending can suffer a sort of positional inconsistency within the word itself. Additionally, these fixed-size splits can be detrimental due to the imbalance of information in the resulting downsampled tokens. Referring back to our example in Table~\ref{tab:ds_methods}, even if tokens are corrected positionally, the word ``great'' is split into two tokens, with the last character getting its own dedicated token while carrying minimal information. 

The Buffered Fixed method offers positional consistency and length consistency, but not morpheme consistency, so comparing this to SDD should tell us the importance of this third factor. 

\section{Architecture} \label{sect:arch}

\begin{figure*}
    \centering
    \includegraphics[scale=0.19]{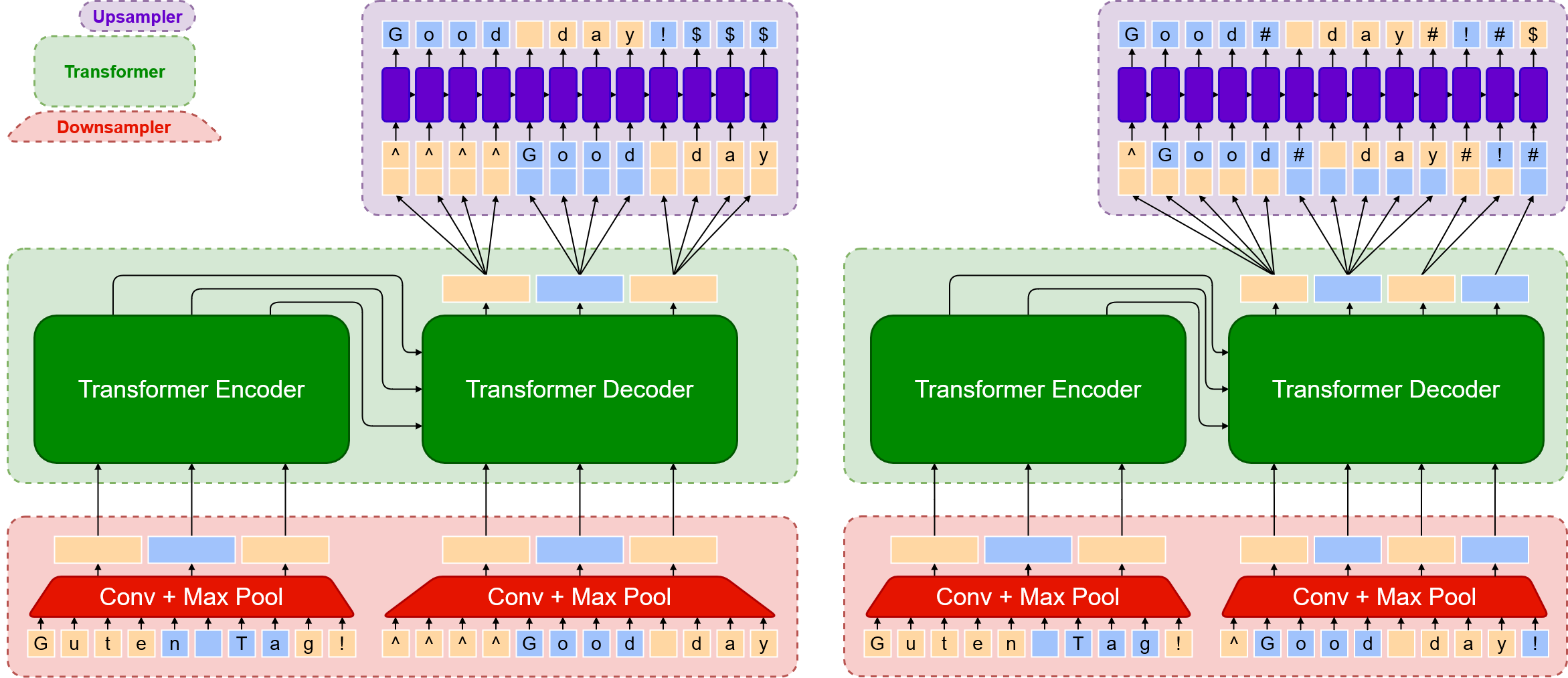}
    \caption{Fixed-size downsampling (left) versus variable-length downsampling (right). Beginning-of-sentence, end-of-sentence, and end-of-word tokens are denoted by \textasciicircum, \$, and \#, respectively.}
    \label{fig:architecture}
\end{figure*}

We now explain the architecture used in our experiments. We build off of the previous work by using the CNN downsampling architecture followed by the Transformer and using \citet{libovicky2021don}'s two-step decoding with an LSTM for upsampling. This previous work was only applied to fixed-length downsampling and upsampling, however the aforementioned WDD and SDD methods require variable-length downsampling and upsampling. Thus, we explain how this is accomplished in the next two subsections. 
Figure \ref{fig:architecture} shows the architectures used in our translation experiments.\footnote{Our encoder-only experiments use the same encoder architecture (including the downsampler) with a linear output layer added.}

\subsection{Variable-length Downsampling}
The downsampling module for WDD and SDD is identical to 
that of the fixed-size model, with the exception that the max pooling is computed over all characters in a word or subword. On the decoder side, we additionally require the lengths of each word or subword token in order to create a causal mask which allows the association of characters within the same block, while preventing the association of characters from future blocks.

\subsection{Variable-length Upsampling}\label{sec:upsampling}
While the downsampling module ensures that the Transformer receives word or subword-level tokens, we still require a method for upsampling back to characters. \citet{libovicky2021don} introduced an effective two-step decoder, consisting of the Transformer followed by an LSTM which takes as input the hidden representation of the Transformer decoder, the character embedding of the previous character, and the previous LSTM hidden state. This has previously only been applied to methods with a fixed-size downsampling, and as such we need to make some modifications to allow it to work with variable-length sequences. 

The top blocks of figure \ref{fig:architecture} show the original and modified versions of the two-step decoder. The input to the LSTM is first modified. In the original case, with a downsampling factor of 4, the hidden representation is repeated 4 times, and each is concatenated with individually learned character embeddings for the block of 4 characters. The LSTM then must predict the next block, which effectively means each character generated is conditioned on the character 4 steps back. 

With the modified version, the hidden representation is repeated the same number of times as the length of the next block plus one, as we add in an end-of-word token for each block to the character embeddings and labels.\footnote{The end-of-word token is added to stop the LSTM's generation of a word at generation time.}
Each hidden representation is concatenated with the character embeddings, shifted over by 1. Although the character embeddings fed to the LSTM are no longer associated with the respective hidden block, since the embeddings are individually learned, no information from future blocks is accessible (thus avoiding the leaking issue described in \citet{edman2022patching}).

Since LSTMs are known to struggle as the lengths of sequences get longer, we also limit the lengths of each subword,\footnote{We use the argument from SentencePiece \texttt{max$\_$sentencepiece$\_$length} to achieve this.} which prevents the joining of subwords beyond a specified character length. This minimizes the length inconsistency previously mentioned in Section~\ref{subsect:len_cons}.

\sisetup{table-format = 2.2, retain-explicit-plus}
\begin{table*}[!htp]\centering
\scriptsize
\sisetup{table-format = 2.3, round-mode=places, round-precision=3, retain-explicit-plus}
\begin{tabular}{lrrrrrrrrrrrrrr}\toprule
&\multicolumn{7}{c}{BLEU} &\multicolumn{7}{c}{COMET} \\\cmidrule(ll){2-8} \cmidrule(ll){9-15}
Method &\multicolumn{1}{c}{de-en} &\multicolumn{1}{c}{en-de} &\multicolumn{1}{c}{ar-en} &\multicolumn{1}{c}{en-ar} &\multicolumn{1}{c}{tr-en} &\multicolumn{1}{c}{en-tr} &\multicolumn{1}{c}{avg} &\multicolumn{1}{c}{de-en} &\multicolumn{1}{c}{en-de} &\multicolumn{1}{c}{ar-en} &\multicolumn{1}{c}{en-ar} &\multicolumn{1}{c}{tr-en} &\multicolumn{1}{c}{en-tr} &\multicolumn{1}{c}{avg} \\\cmidrule(ll){1-8} \cmidrule(ll){9-15}
Fixed &24.32 &21.06 &22.63 &9.24 &12.78 &9.32 &16.56 &0.019 &-0.237 &-0.072 &-0.044 &-0.218 &-0.003 &-0.093 \\
Buf. Fixed &\bfseries 28.71 &24.72 &27.48 &10.69 &14.79 &10.40 &19.46 & 0.234 &\bfseries 0.047& 0.185 & 0.132 &-0.061 &0.137 & 0.112 \\
WDD &27.51 &23.72 &26.71 &11.32 &13.27 &8.92 &18.58 &0.175 &-0.066 &0.151 &0.061 &-0.164 &-0.004 &0.025 \\
SDD &27.78 &\bfseries 25.12 &\bfseries 27.74 &\bfseries 11.95 &\bfseries 15.99 &\bfseries 11.31 &\bfseries 19.98 &\bfseries 0.237 &0.023 &\bfseries 0.213 &\bfseries 0.146 &\bfseries -0.006 &\bfseries 0.175 &\bfseries 0.131 \\ \cmidrule(ll){1-8} \cmidrule(ll){9-15}
Position &+4.39 &+3.66 &+4.84 &+1.45 &+2.01 &+1.08 &+2.90 & +0.214 &+0.284 &+0.258 &+0.176 &+0.157 &+0.139 &+0.205    \\
Length &+0.27 &+1.40 &+1.03 &+0.63 &+2.72 &+2.39 &+1.41 &+0.061 &+0.090 &+0.062 &+0.086 &+0.158 &+0.179 &+0.106 \\
Morpheme &-0.93 &+0.41 &+0.27 &+1.25 &+1.20 &+0.91 &+0.52 &+0.003 &-0.024 &+0.028 &+0.014 &+0.055 &+0.038 &+0.019   \\
\bottomrule
\end{tabular}

\caption{Comparison of the 4 different downsampling methods, with an ablation of positional consistency (Buf. Fixed - Fixed), length consistency (SDD - WDD), and morpheme consistency (SDD - Buf. Fixed) using two evaluation metrics (BLEU and COMET)}.\label{tab:ds_res}
\end{table*}

\section{Experimental Setup} \label{sect:exp}
Our code is made available on GitHub.\footnote{\url{https://github.com/Leukas/SDD}}
We experiment with translation, using the encoder-decoder, as well as two encoder-only tasks: NLI and review classification.\footnote{Review classification is classifying product reviews from 1 to 5 stars based on the review title and content.} While we focus mainly on improving translation, we also include these encoder-only tasks to test the importance of the choice of downsampling method for non-generative tasks.

We compare several models, including the 4 downsampling methods (Fixed, Buffered Fixed, WDD, and SDD) , as well as a subword-level model and a character-level model,\footnote{By character-level we in fact mean byte-level, aligning with previous work. This applies to our downsampling models as well.} both of which use the standard Transformer architecture, requiring no downsampling or upsampling module. 

For translation, we experiment with 3 language pairs: English--Arabic, English--German, and English--Turkish. We chose these language pairs as they exhibit different levels of linguistic similarity and morphological richness. We evaluate our models with BLEU \cite{papineni-etal-2002-bleu} and COMET \cite{rei-etal-2020-comet}.

The full details of the datasets used for all tasks can be found in Appendix~\ref{app:datasets}. To keep our research eco-friendly and to allow for faster iteration on our models, we train our models on smaller translation datasets, consisting of roughly 200 thousand sentence pairs per language pair, and roughly 500 thousand and 5 million sentences for NLI and review classification, respectively. We discuss the potential for these models in the high-resource setting in Section~\ref{sect:discussion}.

For SDD, we change the value of the \verb|max_sentencepiece_length| argument in SentencePiece to achieve an effective downsampling factor as close to our fixed-size downsampling counterparts as possible.\footnote{By default, it is set to 16, which inhibits the performance of the LSTM in the decoder.} To maintain a comparability to a strict downsampling of 4, we set the \verb|max_sentencepiece_length| and the vocabulary size such that the average downsampling factor is around 4.
We achieve this by first setting the vocabulary size to the size recommended by VOLT \cite{xu2020vocabulary}, then lowering the \verb|max_sentencepiece_length| until the average downsampling factor is close to 4.\footnote{The average downsampling factor is calculated by averaging the number of bytes per subword token in the training data.}
We chose 4 as it is roughly equivalent to the ratio of the number of bytes per subword token when comparing the ByT5 and mT5 models across all the languages in the mC4 corpus \cite{xue2022byt5}.

Our parameter setup for models, vocabulary, and training are noted in Appendix~\ref{app:params}.

\section{Results} \label{sect:results}

We start by comparing the 4 downsampling methods, discerning the importance of the positional, length, and morpheme consistency. 
We then compare our best-performing downsampling method to our two standard baselines: the subword-level and character-level models. 
Third, we compare these models in their ability to generalize to out-of-domain datasets. 
Lastly, we look at the encoder-only tasks, comparing all methods thus far.

\begin{table*}[!htp]\centering
\scriptsize
\sisetup{table-format = 2.3, round-mode=places, round-precision=3, retain-explicit-plus}
\begin{tabular}{lrrrrrrrrrrrrrr}\toprule
&\multicolumn{7}{c}{BLEU} &\multicolumn{7}{c}{COMET} \\\cmidrule(ll){2-8} \cmidrule(ll){9-15}
Method &\multicolumn{1}{c}{de-en} &\multicolumn{1}{c}{en-de} &\multicolumn{1}{c}{ar-en} &\multicolumn{1}{c}{en-ar} &\multicolumn{1}{c}{tr-en} &\multicolumn{1}{c}{en-tr} &\multicolumn{1}{c}{avg} &\multicolumn{1}{c}{de-en} &\multicolumn{1}{c}{en-de} &\multicolumn{1}{c}{ar-en} &\multicolumn{1}{c}{en-ar} &\multicolumn{1}{c}{tr-en} &\multicolumn{1}{c}{en-tr} &\multicolumn{1}{c}{avg} \\\cmidrule(ll){1-8} \cmidrule(ll){9-15}
Subword &27.23 &24.08 &25.59 &11.22 &14.73 &11.30 &19.02 &0.203 &\bfseries 0.073 &0.063 &0.120 &-0.097 &0.176 &0.090 \\
Char &27.37 &24.32 &26.34 &\cellcolor[HTML]{e06666}8.73 &\cellcolor[HTML]{add8e6}15.52 &\cellcolor[HTML]{add8e6}\bfseries 11.87 &19.03 &0.222 &0.040 &\cellcolor[HTML]{add8e6}0.142 &\cellcolor[HTML]{e06666}-0.008 &\cellcolor[HTML]{add8e6}-0.026 &\bfseries 0.190 &
0.093 \\
SDD &\bfseries 27.78 &\cellcolor[HTML]{add8e6}\bfseries 25.12 &\cellcolor[HTML]{add8e6}\bfseries 27.74 &\cellcolor[HTML]{add8e6}\bfseries 11.95 &\cellcolor[HTML]{add8e6}\bfseries 15.99 &11.31 &\bfseries 19.98 &\cellcolor[HTML]{add8e6}\bfseries 0.237 &\cellcolor[HTML]{e06666}0.023 &\cellcolor[HTML]{add8e6}\bfseries 0.213 &\bfseries 0.146 &\cellcolor[HTML]{add8e6}\bfseries -0.006 &0.175 &\bfseries 0.131\\
\bottomrule
\end{tabular}


\caption{Translation results of traditional subword models and character models without downsampling compared to character models with subword-delimited downsampling (SDD). {\sethlcolor{hlblue}\hl{Blue}} and {\sethlcolor{hlred}\hl{red}} denote a significant positive or negative difference (p < 0.05) with respect to the Subword model.  }\label{tab:nmt_bleu}
\end{table*}

\subsection{Comparing Downsampling Methods}
First comparing the various downsampling methods Table~\ref{tab:ds_res}, we see that as expected from Section~\ref{sect:downsampling} our novel SDD method performs best overall. 

Of the three factors, positional consistency appears most important, making up the greatest portion of the performance increase. Given that positional inconsistency means there is often no subword-like structure to the downsampled tokens, it is understandable that the model has the most difficulty translating in this scenario.

Length consistency contributes some improvement as well, showing that there is indeed a bottleneck effect when downsampling longer words, and that there is merit in the model splitting such words into multiple tokens prior to the Transformer. 

Morpheme consistency is least important; however it seems to have a larger impact on the more morphological languages of Arabic and Turkish. This gives evidence towards the idea that the subword splitting of SentencePiece is likely more conducive to translating morphemes than translating 4-character chunks. 
Although we use SentencePiece due to its ubiquity, there are arguably better subword tokenizers with respect to adherence to morphology, such as LMVR \cite{ataman2017linguistically}. These tokenizers are fully compatible with SDD, and they may further increase the disparity we see in morpheme consistency. 


\subsection{Comparing to Subword and Character Models}


We now compare the performance of SDD to subword and character models in Table~\ref{tab:nmt_bleu}. 

Overall, SDD outperforms both subword and character models on both BLEU and COMET. It performs significantly better than the subword model in 4 and 3 cases according to BLEU and COMET, respectively. Only in 1 case in terms of COMET does SDD lead to significantly worse performance.

The Arabic--English language pair shows the largest improvement over the baselines. The reason for this is unclear, though it is the only language pair with an entirely different character set between the source and target (save for numbers, symbols, and some proper nouns).

As SDD was originally intended to shore up the weaknesses of the previous downsampling methods, the aim was to perform on par with the subword and character models. The slight increase in performance shows that there is some benefit in using a combination of the two, namely using character-level input while operating on a subword-level within the Transformer encoder-decoder. 





\subsection{Out-of-domain Generalization} \label{subsect:ood}
\citet{boukkouri2020characterbert} found that with CharacterBERT, the models generalized better to out-of-domain encoder-only tasks such as classification of medical data. They argue that because CharacterBERT does not have a strict vocabulary, and it learns more general properties of language which can be useful for unseen words. As we see evidence of the same in our embedding analysis (see Section~\ref{sect:analysis}), we similarly test our models on two out-of-domain translation datasets. For all languages, we test on FLoRes \cite{goyal2022flores}, which consists of Wikipedia data. For English--German, we additionally test on the WMT21 Biomedical Shared Task test set.

The FLoRes and biomedical results are shown in Tables \ref{tab:flores} and \ref{tab:biomed}, respectively. Here we see the SDD model and character model both outperforming the subword model, as expected. The results on FLoRes favor the SDD model, while the biomedical results favor the character model. The biomedical data is arguably ``more out-of-domain'' than FLoRes, since it contains a large amount of medical terminology unlikely to appear in the training data. This may indicate that the character downsampling is somewhat sensitive to the vocabulary it is trained on, and as such it generalises better than the subword model but not as well as the character model.

Since the SDD model only uses a subword vocabulary to determine the lengths for downsampling, it is perhaps possible to use a different vocabulary when switching domains, namely one that offers better downsampling for the out-of-domain words. We leave this for future research.

\begin{table*}[!htp]\centering
\scriptsize


\sisetup{table-format = 2.3, round-mode=places, round-precision=3, retain-explicit-plus}
\begin{tabular}{lrrrrrrrrrrrrrr}\toprule
&\multicolumn{7}{c}{BLEU} &\multicolumn{7}{c}{COMET} \\\cmidrule(ll){2-8} \cmidrule(ll){9-15}
Method &\multicolumn{1}{c}{de-en} &\multicolumn{1}{c}{en-de} &\multicolumn{1}{c}{ar-en} &\multicolumn{1}{c}{en-ar} &\multicolumn{1}{c}{tr-en} &\multicolumn{1}{c}{en-tr} &\multicolumn{1}{c}{avg} &\multicolumn{1}{c}{de-en} &\multicolumn{1}{c}{en-de} &\multicolumn{1}{c}{ar-en} &\multicolumn{1}{c}{en-ar} &\multicolumn{1}{c}{tr-en} &\multicolumn{1}{c}{en-tr} &\multicolumn{1}{c}{avg} \\\cmidrule(ll){1-8} \cmidrule(ll){9-15}
Subword &19.88 &16.79 &13.90 &9.07 &16.25 &11.02 &14.49 &-0.158 &-0.385 &-0.344 &-0.292 &-0.053 &-0.031 &-0.210 \\
Char &\bfseries 21.97 &\bfseries 17.46 &15.86 &8.95 &17.59 &\bfseries 12.03 &15.64 &\bfseries 0.003 &\bfseries -0.297 &-0.132 &-0.340 &0.065 &\bfseries 0.049 &-0.109 \\
SDD &21.45 &\bfseries 17.46 &\bfseries 16.75 &\bfseries 10.73 &\bfseries 17.85 &12.01 &\bfseries 16.04 &-0.065 &-0.330 &\bfseries -0.094 &\bfseries -0.154 &\bfseries 0.074 &0.035 &\bfseries -0.089 \\
\bottomrule
\end{tabular}

\caption{Results on FLoRes evaluation set.}\label{tab:flores}
\end{table*}

\begin{table}[!htp]\centering

\scriptsize
\sisetup{table-format = 2.3, round-mode=places, round-precision=3, retain-explicit-plus}
\begin{tabular}{lrrrr}\toprule
&\multicolumn{2}{c}{BLEU} &\multicolumn{2}{c}{COMET} \\\cmidrule{2-3} \cmidrule(ll){4-5}
Method &\multicolumn{1}{c}{de-en} &\multicolumn{1}{c}{en-de}&\multicolumn{1}{c}{de-en} &\multicolumn{1}{c}{en-de} \\\cmidrule{1-3} \cmidrule(ll){4-5}
Subword &9.37 &4.36 &-0.729 &-1.119 \\
Char &\bfseries 11.71 &\bfseries 4.74&\bfseries -0.499 &\bfseries -0.947 \\
SDD &9.41 &4.05 &-0.659 &-1.036 \\
\bottomrule
\end{tabular}
\caption{Results on Biomedical evaluation set.}\label{tab:biomed}
\end{table}

\subsection{Encoder-only Tasks}
The accuracies achieved in the encoder-only tasks are shown in Table~\ref{tab:enc_res}. The character-level model has surprisingly low accuracies. Given that the main difference between the character model and the other models is the longer sequence length fed into the Transformer, we expect the complexity of the self-attention patterns necessary for these tasks is more difficult to learn when data is limited. 

\begin{table}[!htp]\centering
\scriptsize
\begin{tabular}{lrrrr}\toprule
Method &NLI (\%) &RC (\%) \\ \midrule
Subword &80.14 &73.90 \\ 
Char &60.75 &68.86 \\ 
Fixed &81.01 &73.28 \\ 
Buf. Fixed &81.81 &73.05 \\ 
WDD &\bfseries 81.89 &73.69 \\ 
SDD &81.30 &\bfseries 74.22 \\ 
\bottomrule
\end{tabular}
\caption{Results of NLI and review classification (RC).}
\label{tab:enc_res}
\end{table}

Our SDD method outperforms both subword and character baselines on both tasks.
Unlike with translation, the downsampling models show little difference in performance. As SDD is intended to help by providing a consistent tokenization, it seems that this consistency is less important for sequence classification tasks. This is probably because there is no need for character or word recovery: the model does not need to reconstruct any of its input, so it can potentially lose some character information while still learning to correctly classify.

A token classification task such as part-of-speech tagging may be more difficult for the fixed-size downsampling model, although it is not clear how to apply such a model to a token classification task, given the mismatch in downsampling blocks and token labels. The SDD model can however be applied to token classification in the same manner a subword model would be applied.

Another reason for there being more of a gap in the performance for translation might be that the cross-attention is what benefits most from the three token consistency factors. Inconsistent tokens on both the source and target side likely make learning what to attend to quite difficult. 




\section{Embedding Analysis} \label{sect:analysis}
\begin{table*}[!htp]\centering
\scriptsize
\begin{tabular}{lrrrrrr}\toprule
& &Grammatical &Close Spell  &Far Spell &Far Synonym \\ \midrule
Subword &Seen &1.03 ± 0.08 &0.17 ± 0.06 &0.05 ± 0.05 &0.18 ± 0.05 \\
$\mu$: 0.01 &Half-seen &-0.09 ± 0.07 &-0.09 ± 0.02 &-0.12 ± 0.02 &-0.12 ± 0.02 \\
$\sigma$: 0.08 &Unseen &-0.12 ± 0.12 &-0.10 ± 0.05 &-0.10 ± 0.02 &-0.06 ± 0.09 \\ \midrule
SDD &Seen &5.75 ± 0.05 &4.68 ± 0.06 &-0.52 ± 0.04 &-0.49 ± 0.04 \\
$\mu$: 0.31 &Half-seen &6.09 ± 0.09 &5.17 ± 0.06 &-0.43 ± 0.04 &-0.40 ± 0.04 \\
$\sigma$: 0.09 &Unseen &6.06 ± 0.14 &5.62 ± 0.11 &-0.27 ± 0.04 &-0.21 ± 0.13 \\
\bottomrule
\end{tabular}

\caption{Model average $z$-scores for the respective test sets, with a 95\% confidence interval.}\label{tab:z_scores}
\end{table*}
The sole difference in the architecture of the encoder-only models is their embeddings, or the hidden representations prior to being fed into the Transformer. As such, we extract these embeddings for words to analyze their differences. 
We compare the models in their word embedding similarity (i.e. cosine similarity) for pairs of words. We generate 4 test sets: 
\begin{enumerate}
    \setlength\itemsep{-0.2em}
    \item Grammatical pairs - Pairs which share a common lemma. 
    \item Close pairs - Pairs with a Levenshtein distance of 1.
    \item Far pairs - Pairs with a max Levenshtein distance. 
    \item Far Synonyms - Synonyms with a max Levenshtein distance.
\end{enumerate}
The first two sets are meant to distinguish beneficial character-level similarity (e.g. ``take'' and ``takes'') from detrimental similarity (e.g. ``pour'' and ``tour''). The latter two sets are meant to distinguish the opposite, when a lack of similarity may be appropriate (e.g. ``bulb'' and ``faster'') or inappropriate (e.g. ``camp'' and ``tent''). We limit the vocabulary used in the test sets to words in the subword vocabulary that only require a single token. We also filter out tokens which have no synonyms, effectively also removing non-words and non-English words. Lemmas and synonyms are gathered using WordNet \cite{miller1995wordnet}. 

We split the test sets additionally into 3 sub-categories: ``seen'', where both words in the pair have been seen during training, ``half-seen'', where only one word is seen in training, and ``unseen'', where neither is seen in training.\footnote{For these experiments, we used separate models trained using the same initial vocabulary as T5-Small, in order to get a larger number of single-token words to evaluate on. This is why there exist words in the vocabulary that are not seen during training. The performance of these models is similar to those reported in Table~\ref{tab:enc_res}.} The sizes of each split are shown in Table~\ref{tab:emb_data_size} in Appendix~\ref{app:datasets}. The results are shown in Table~\ref{tab:z_scores}.

As expected, the subword model has near 0 similarity for half-seen and unseen words since it has no mechanism for developing their embeddings during training. It does recognize grammatical seen words as similar, but it has a more muted response to seen words with similar spelling. In contrast, SDD has a strong response to both seen and unseen words, as the character models can develop embeddings for unseen words according to character-level patterns. We see that both models appear capable of distinguishing between grammatical pairs and pairs with similar spelling but not necessarily similar meaning, though only in the case where both words are seen for the subword model.

There is a stark difference between the z-scores of the Close Spell and Far Spell sets for the SDD model. As for the Far Synonyms, there is a small difference for the subword model, and none for the SDD model. It is possible that these words are distinguished in later layers of the model.



\section{Practicality Discussion} \label{sect:discussion}
While this paper takes a more theoretical approach, using smaller, domain-specific datasets for training NMT, the practical usage is worth considering. Given the prior work of ByT5, Charformer, and CharacterBERT, the consensus appears that with large, pretrained models using character information, the performance on standard metrics is similar, and the outperformance is on out-of-domain data \cite{boukkouri2020characterbert} or data with character-level corrupted input \cite{xue2022byt5}. As such, we expect the same is true for SDD, however the upsampling module used (based on~\citet{libovicky2021don}) appears to be a limiting factor. We explain this in further detail in Appendix~\ref{app:wmt}, where we train and test the models on the larger WMT14 German--English dataset, finding that the performance of the subword model deteriorates when the same two-step decoding method is added.

Conversely, we expect such models to be useful in lower-resource settings. In such an scenarios, we show that the inclusion of character-level information improves performance beyond that of subword models (Appendix~\ref{app:xhzu}).

\begin{table}[!htp]\centering
\scriptsize
\begin{tabular}{lrrrrrrr}\toprule
&\multicolumn{3}{c}{Translation} &\multicolumn{3}{c}{Encoder Tasks} \\\cmidrule(lr){2-4} \cmidrule(lr){5-7}
&\multicolumn{1}{c}{Iter} &\multicolumn{1}{c}{Epochs} &\multicolumn{1}{c}{Total} &\multicolumn{1}{c}{Iter} &\multicolumn{1}{c}{Epochs} &\multicolumn{1}{c}{Total} \\\midrule
Char &$\times3.58 $ &$\times1.31 $ &$\times4.69 $ &$\times2.23 $ &$\times2.04 $ &$\times4.55$ \\
SDD &$\times2.61$ &$\times1.28 $ &$\times3.34 $ &$\times1.34 $ &$\times1.36 $ &$\times1.82$ \\
\bottomrule
\end{tabular}
\caption{Training time ratios with respect to the subword model. ``Iter'' refers to the iteration speed (e.g. Char is 3.58 times slower than the subword model per iteration for translation), ``Epochs'' refers to the total number of epochs needed, and ``Total'' is the product of the first two.}\label{tab:times}
\end{table}

In terms of training time, we report the training time ratios with respect to the subword model in Table~\ref{tab:times}.\footnote{All experiments were conducted on a single Nvidia V100 GPU.} While the SDD model is considerably slower than the subword model in terms of training time, it still performs better than the character model, particularly for encoder-only tasks, suggesting that decoding is the main source of the slower training times.

One potential use of the SDD model that is yet unexplored comes in the form of adapting existing subword models to take character input. Since pretraining is a costly endeavor, and since there are beneficial characteristics of character-level models such as out-of-domain generalization, it may be possible to improve a subword model by adapting it to use the SDD downsampling module rather than its own subword embeddings. Previous work in adaptation has shown success in adapting models to different tasks \cite{ustun2020udapter,pfeiffer2020mad} and languages \cite{bapna2019simple,artetxe2019cross,ustun2021multilingual}, so a similar approach may be useful here. 

\section{Conclusion} \label{sect:conc}
Previous work has casted doubt on the usefulness of character-level NMT models, due to their lack of improvement over subword models despite using more fine-grained information while being also slower. Downsampling modules added to the character models have previously been proposed but have always come at the cost of accuracy. 

We show that it is possible to downsample without sacrificing accuracy, by downsampling based on the lengths of subwords. This novel downsampling outperforms the previous downsampling methods, as well as character and subword models on the majority of language pairs tested. 

There are several avenues for future research. While much work has been done on optimizing a vocabulary for a subword model, finding the optimal lengths for subword-delimited downsampling is still an open problem. The most promising may in fact be adapting any of the numerous pretrained subword models to use characters as input. Overall, character-level models show promise that has yet to be fully realized.

\section{Limitations} \label{sect:limit}
As we mention in Sections \ref{sect:exp} and \ref{sect:discussion}, our largest limitation is that the majority of our testing is done on smaller datasets, consisting of roughly 200 thousand sentence pairs per language pair. While we do test on a larger dataset for German$\rightarrow$English, it is limited to only a single language pair, and single direction, so it is still an open question whether our method works well in general in higher-resource settings.

Additionally, we only test on 3 language pairs in our main results, all of which have English on the source or target side. It is possible that our method only works well in circumstances where English is present in the language pair, or it only works well where the other language is either German, Arabic, or Turkish. Notably, languages such as Chinese can have characters that hold the same meaning as a word in English, and as such subword tokenizers like SentencePiece may be less useful. Subsequently, SDD may be less useful for these types of languages. 

In terms of the tasks tested, we mainly focus on translation, so we can make no claims about the performance on other generative tasks. We test on two discriminative tasks, NLI, and review classification, however both are tested without following the popular pretrain-then-finetune paradigm, making the results difficult to compare to existing work. The scope of these tests is also limited to English only. 

In terms of parameters tested, we mainly follow previous work, so it is possible that our method does not perform as well (or possibly performs better) under different parameter settings. To keep our carbon impact minimal, we opted for using VOLT to determine an optimal vocabulary size (which requires no training of NMT models), rather than the standard grid search approach. VOLT does not guarantee an optimal vocabulary size however, and this may have an impact on our results, be it favorably or unfavorably. 

Finally, as we note in Appendix~\ref{subsect:app_train}, our method is considerably slower than the subword model. However, it is also considerably faster than the character model while also performing better on the majority of the tasks tested. As this work is exploring new territory, it is very likely that our implementation is not as efficient as it could be. 



\section{Acknowledgements}
We thank the Center for Information Technology of the University of Groningen for their support and for providing access to the Peregrine high performance computing cluster.

\bibliography{anthology,custom}
\bibliographystyle{acl_natbib}

\appendix
\section{Datasets} \label{app:datasets}

\begin{table*}[!htp]\centering

\scriptsize
\begin{tabular}{lrrrrrr}\toprule
& \multicolumn{3}{c}{Main} & \multicolumn{1}{c}{Low-Resource} & \multicolumn{1}{c}{High-Resource} \\ \cmidrule(lr){2-4} \cmidrule(lr){5-5} \cmidrule(lr){6-6}
Lang pair &\multicolumn{1}{c}{en$\leftrightarrow$de} &\multicolumn{1}{c}{en$\leftrightarrow$ar} &\multicolumn{1}{c}{en$\leftrightarrow$tr} &\multicolumn{1}{c}{xh$\leftrightarrow$zu} & \multicolumn{1}{c}{de$\rightarrow$en} \\ \midrule
Train &IWSLT2017 &IWSLT2017 &SETIMES &CC-Aligned & WMT14 \\
Size (k) &199 &224 &197 &85 & 4509 \\
Eval &IWSLT2017 &IWSLT2017 &WMT17/18 dev &WMT21 dev & WMT13/14 dev \\ 
Domain & TED Talks & TED Talks & News & Crawl & News / Crawl / Govt \\
\bottomrule
\end{tabular}
\caption{Datasets used for translation, showing the names of the training and evaluation sets, the size of the training data (in sentence pairs), and the domain. }\label{tab:nmt_data}
\end{table*}

\begin{table}[!htp]\centering

\scriptsize
\begin{tabular}{lrr}\toprule
Task & NLI & RC \\\midrule
Train &SNLI train &Books\_v1\_00 \\
Size (k) &547 &5711 \\
Eval &SNLI dev/test & Books\_v1\_00 / 02 \\
Domain &Image captions &Book reviews \\
\bottomrule
\end{tabular}
\caption{Datasets used for encoder-only tasks, showing the names of the training and evaluation sets, the size of the training data (in premise/hypothesis pairs or reviews), and the domain.}\label{tab:enc_data}
\end{table}

The datasets used for translation are shown in Table~\ref{tab:nmt_data}. In summary, we use IWSLT2017,\footnote{\label{note:iwslt}\url{https://sites.google.com/site/iwsltevaluation2017/TED-tasks}} SETIMES,\footnote{\url{https://opus.nlpl.eu/SETIMES.php}} and datasets provided by WMT shared tasks.\footnote{\url{https://statmt.org/wmt21/translation-task.html}} 

The datasets used for NLI and review classification (RC) are shown in Table~\ref{tab:enc_data}.
For NLI, we use the SNLI corpus \cite{bowman2015large}, and for RC we use the Amazon Customer Reviews Dataset.\footnote{\url{https://s3.amazonaws.com/amazon-reviews-pds/readme.html}} For SNLI we use the predefined train/dev/test splits, and for RC we take a 99/1 split of the \verb|Books_v1_00| subset for training and validation, and use 1\% of the \verb|Books_v1_02| subset for testing (amounting to roughly 22k test sentences).

Concerning the evaluation sets we generated for Section~\ref{sect:analysis}, we show the sizes of the generated sets in Table~\ref{tab:emb_data_size}. We limited the size of the test sets to 2000 word pairs, randomly selecting from those generated.

\begin{table}[!htp]\centering

\scriptsize
\begin{tabular}{lrrrrr}\toprule
&Grammatical &Close spell &Far spell &Far Synonym \\ \midrule
Seen &1873 &2000 &2000 &2000 \\
Half-seen &273 &2000 &2000 &2000 \\
Unseen &92 &525 &2000 &164 \\
\bottomrule
\end{tabular}
\caption{Size of generated datasets for evaluating the embeddings of word pairs.}\label{tab:emb_data_size}
\end{table}

\section{Parameters} \label{app:params}
\subsection{Architecture}
For the CNN downsampling module, we use the same parameters as \citet{libovicky2021don}. 

We use Transformer Base for our Encoder-Decoder, only modifying the maximum position embeddings to 2048 to account for the increased length of our character models.

For the upsampling module, we use the same parameters as used in \citet{libovicky2021don}, with the exception of the linear layer which projects the hidden representations of the Transformer decoder to the LSTM. The output size used in \citet{libovicky2021don} is 64 times the downsampling factor. For variable-length downsampling, this needs to be set to the maximum word length in bytes plus one (accounting for the end-of-word token). 

\begin{table}[!htp]\centering
\scriptsize
\begin{tabular}{lrrrrr}\toprule
&de$\leftrightarrow$en &ar$\leftrightarrow$en &tr$\leftrightarrow$en &xh$\leftrightarrow$zu \\\midrule
Subword &62M &55M &51M &50M \\
Char &46M &46M &46M &46M \\
SDD &56M &56M &56M &56M \\
\bottomrule
\end{tabular}
\caption{Number of parameters (in millions) for each model trained.}\label{tab:param_count}
\end{table}

The total number of parameters in each model is shown in Table \ref{tab:param_count}. We opted to keep the number of parameters in the Transformer fixed, meaning the variation in parameter count comes only from the embeddings as well as the downsampling and upsampling modules, if they are present. The number of parameters varies across language for the subword model due to the varying size of the vocabulary. 

\subsection{Vocabulary}
\begin{table}[!htp]\centering
\scriptsize
\begin{tabular}{lrrrrrr}\toprule
&\multicolumn{3}{c}{Main} & \multicolumn{1}{c}{LR} & \multicolumn{1}{c}{HR} \\\cmidrule(lr){2-4} \cmidrule(lr){5-5} \cmidrule(lr){6-6}
Lang pair &\multicolumn{1}{c}{en$\leftrightarrow$de} &\multicolumn{1}{c}{en$\leftrightarrow$ar} &\multicolumn{1}{c}{en$\leftrightarrow$tr} &\multicolumn{1}{c}{xh$\leftrightarrow$zu}
&\multicolumn{1}{c}{de$\rightarrow$en}\\\midrule
Vocab size (k) &16 &9 &5 &4 & 16 \\
Max token len &6 &4 &6 &5 & 6 \\
\bottomrule
\end{tabular}
\caption{Vocabulary sizes and max token lengths for the languages tested.}\label{tab:vocab}
\end{table}

As noted in Section~\ref{sect:exp}, we use VOLT to determine the optimal vocabulary sizes. We specify these, along with the max token length that most closely matches an average downsampling factor of 4 in Table~\ref{tab:vocab}. It should be noted that the max token length argument in SentencePiece is determined by the maximum character length, however SDD operates on the byte level. So for English--Arabic, although the max token length is set to 4, the average downsampling factor is in fact above 4, due to the length of the Arabic characters in UTF-8 being typically 2-3 bytes.

\subsection{Training} \label{subsect:app_train}
We use an effective batch size of 128, accumulated over 4 iterations. We use the AdamW optimizer \cite{loshchilov2017decoupled}, with betas 0.9 and 0.999, a learning rate of 2e-4, and a linear scheduler with a warmup of 10000 steps. We apply an early stopping with a patience of 10, using the BLEU score on the validation set as the stopping criterion.

\section{Higher-Resource Translation} \label{app:wmt}

We also run experiments in the higher resource setting of the WMT14 German$\rightarrow$English data. 
Following the principles of \citet{https://doi.org/10.48550/arxiv.1806.00187}, we use larger batch sizes of 50k and 240k tokens for the subword and character models, respectively, both of which average to about 2000 sentences per batch. We also increase the learning rate to 5e-4 and use mixed-precision for the training. 
We include an additional model, which is the subword model with the two-step decoding as used in the models with downsampling. In other words, the subword model is given the same LSTM upsampling head, but the upsampling is simply 1-to-1. We test this to ablate any effect of the upsampler on the results.\footnote{We do not include this in our main results because we did not see a noticeable difference in the performances there.}

\begin{table}[!htp]\centering
\scriptsize
\sisetup{table-format = 2.3, round-mode=places, round-precision=3, retain-explicit-plus}
\begin{tabular}{lrr}\toprule
&\multicolumn{1}{c}{BLEU} &\multicolumn{1}{c}{COMET}\\\midrule
Subword &\bfseries 28.62 & \bfseries 0.431\\
Two-step subword &27.49 &0.360 \\
SDD &27.27 &0.364 \\
\bottomrule
\end{tabular}
\caption{Translation results on the WMT14 DE$\rightarrow$EN dataset.}\label{tab:wmt14}
\end{table}

Our results are shown in Table~\ref{tab:wmt14}.
We can see that the standard subword model performs best, however the addition of the two-step decoder hurts performance significantly. The subword model with the two-step decoder still performs on par with our SDD model, confirming our expectations. We conclude that SDD is competitive in higher-resource settings, since it achieves similar scores to the subword variant that also has a two-step decoder. This two-step decoder, while effective on lower-resource settings, does not scale well to higher-resource settings without any modification. This raises the question as to whether an upsampling method exists that has better scaling ability. We leave this for future research to explore.

\section{Lower-Resource Translation} \label{app:xhzu}
To analyze our models on lower-resource translation, we chose to train and evaluate on Xhosa--Zulu, specifically the data provided by the WMT2021 Shared Task. Typically, in lower-resource scenarios, other techniques such as multilingual transfer learning and back-translation are applied to improve the models. We chose this language pair as the performance gain from employing these techniques is less substantial \cite{wei2021hw}.

\begin{table}[!htp]\centering

\scriptsize
\begin{tabular}{lrrrrr}\toprule
&\multicolumn{2}{c}{\hspace{1em}BLEU} &\multicolumn{2}{c}{COMET} \\\cmidrule(ll){2-3} \cmidrule(ll){4-5}
Method &xh-zu &zu-xh &xh-zu &zu-xh \\\cmidrule{1-3} \cmidrule(ll){4-5}
Subword &4.59 &5.89 &-0.322 &-0.115 \\
Char &\bfseries 6.51 &\bfseries 6.42 &\bfseries -0.007 &\bfseries 0.029 \\
SDD &6.35 &6.32 &-0.128 &-0.029 \\
\bottomrule
\end{tabular}
\caption{Translation Results for Xhosa--Zulu.}\label{tab:xhzu}
\end{table}

The results show that the character-level model without downsampling performs best, with SDD as a close second. Xhosa and Zulu, being closely related languages, likely would benefit greatly from character-level translation regardless of the amount of data, given that many of the differences between the two are on the character level. Of course, more thoroughly evaluating SDD on low-resource settings would require a pretrain-then-finetune approach, which we reserve for future research.

\end{document}